\documentclass[conference]{IEEEtran}
\ifCLASSINFOpdf
\else
\fi

\usepackage{amsmath,graphicx,amssymb,amsthm,amsmath,mathtools}

\usepackage{subfigure,epstopdf,algorithm}
\usepackage{array}
\usepackage{xcolor}
\usepackage{thmtools}

\usepackage{times}
\usepackage{eqparbox}

\usepackage[noend]{algpseudocode}

\def\x{{\mathbf x}}

\def\t{{\mathbf t}}

\def\h{{\mathbf h}}
\def\H{{\mathbf H}}
\def\T{{\mathbf T}}

\begin{document}
%
\title{Extreme Learning Machine for Graph Signal Processing}

\author{\IEEEauthorblockN{Arun~Venkitaraman, Saikat Chatterjee, Peter H{\"a}ndel}
\IEEEauthorblockA{Department of Information Science and Engineering \\                    
	School of Electrical Engineering, KTH Royal Institute of Technology, Stockholm, Sweden                 \\
	arunv@kth.se, sach@kth.se, ph@kth.se}
}


%


\maketitle

\begin{abstract}
In this article, we improve extreme learning machines for regression tasks using a graph signal processing based regularization. We assume that the target signal for prediction or regression is a graph signal. With this assumption, we use the regularization to enforce that the output of an extreme learning machine is  smooth over a given graph. Simulation results with real data confirm that such regularization helps significantly when the available training data is limited in size and corrupted by noise. 
\end{abstract}


%
\IEEEpeerreviewmaketitle

\section{Introduction}
Extreme learning machines (ELMs) have emerged as an active area of research within the machine learning community \cite{elm_Huang2012}. 
ELMs differ from the traditional approaches such as neural networks (NNs) and suppor vector machines (SVMs) in an important respect: the parameters of hidden nodes of the ELM are randomly generated and the learning takes place only at the output layer or at the extreme layer by solving a regularized least-squares problem \cite{elm_HUANG2010}. As a result, ELM does not suffer from computational issues that often affect traditional approaches, making the ELM a fast and effective learning approach. ELM despite its simplicity has been shown to have high quality performances in classification and regression tasks, often giving similar or better results in comparison with NNs and SVMs, while being orders of magnitude faster\cite{elm_HUANG2010,elm_HUANG2015,elm_MOHAMMED20112588,elm_8085130}. In fact, the ELM enjoys universal approximation properties: given mild conditions on the activation functions, the ELM can be shown to approximate any continuous(in the case of regression) or piecewise continuous (in the case of classification) target function as the number of neurons in the hidden layer tends to infinity \cite{elm_Huang2006, elm_Huang2012}.
Though traditionally ELMs have been developed as single-layer feed forward networks, multi-layer, distributed, and incremental extensions have also been developed \cite{ elm_LUO2017,elm_multilayer,elm_8057769}.

As with any machine learning paradigm, the performance of the ELM depends on the nature of the training and testing data. The more abundant and reliable the training data be, the better the classification or regression performance be. However, in many applications the training data may be scarce and corrupted by noise. In such cases, it is important to incorporate additional structures during the training phase. In this article, we propose the use of the emerging notion of graph signal processing \cite{Shuman,Sandry2} to enhance the prediction performance of the ELM in regression tasks. In particular, we consider the target or output of the ELM to be smooth signals over a graph, and propose ELM for graph signal processing (ELMG). Our hypothesis is that such an approach results in improved prediction performance. Experiments with real-world data show the validity of our hypothesis. ELMs and graph signal processing have both emerged as directions of much interest in the respective communities, and we believe that our work is a step towards bringing them together.

\section{Preliminaries}
\subsection{Graph signal processing}
Consider a graph of $M$ nodes denoted by $\mathcal{G}=(\mathcal{V},\mathcal{E},\mathbf{A})$ where $\mathcal{V}$ denotes the node set, $\mathcal{E}$ the edge set, and $\mathbf{A}=[a_{ij}]$ the adjacency matrix, $a_{ij}\geq0$. Since we consider only undirected graphs, we have that $\mathbf{A}$ is symmetric \cite{Chung}. A vector $\mathbf{y}=[y(1) y(2) \cdots y(M)]^\top\in\mathbb{R}^{M}$ is said to be a graph signal over $\mathcal{G}$ if $y(m)$ denotes the value of the signal at the $m$th node of the graph \cite{Shuman,Sandry2, Sandry1,Sandry3,chen1,chen2}. The smoothness of a graph signal $\mathbf{y}$ is measured in terms of the quadratic form:
\begin{equation*}
l(\mathbf{y})=\mathbf{y}^\top\mathbf{L}\mathbf{y}=\sum_{(i,j)\in\mathcal{E}}a_{ij}(y(i)-y(j))^2,
\end{equation*}
where
$\mathbf{L}=\mathbf{D}-\mathbf{A}$ is the graph Laplacian matrix, 
$\mathbf{D}$ being the diagonal degree matrix with $i$th diagonal given by $d_{i}=\sum_ja_{ij}$. 
$l(\mathbf{y})$ is a measure of variation of $\mathbf{y}$ across connected nodes: the smaller the value of $l(\mathbf{y)}$ implies the smoother the signal $\mathbf{y}$. 

\subsection{Extreme learning machine}
\begin{figure}[t]
	\centering
	\includegraphics[width=2.8in]{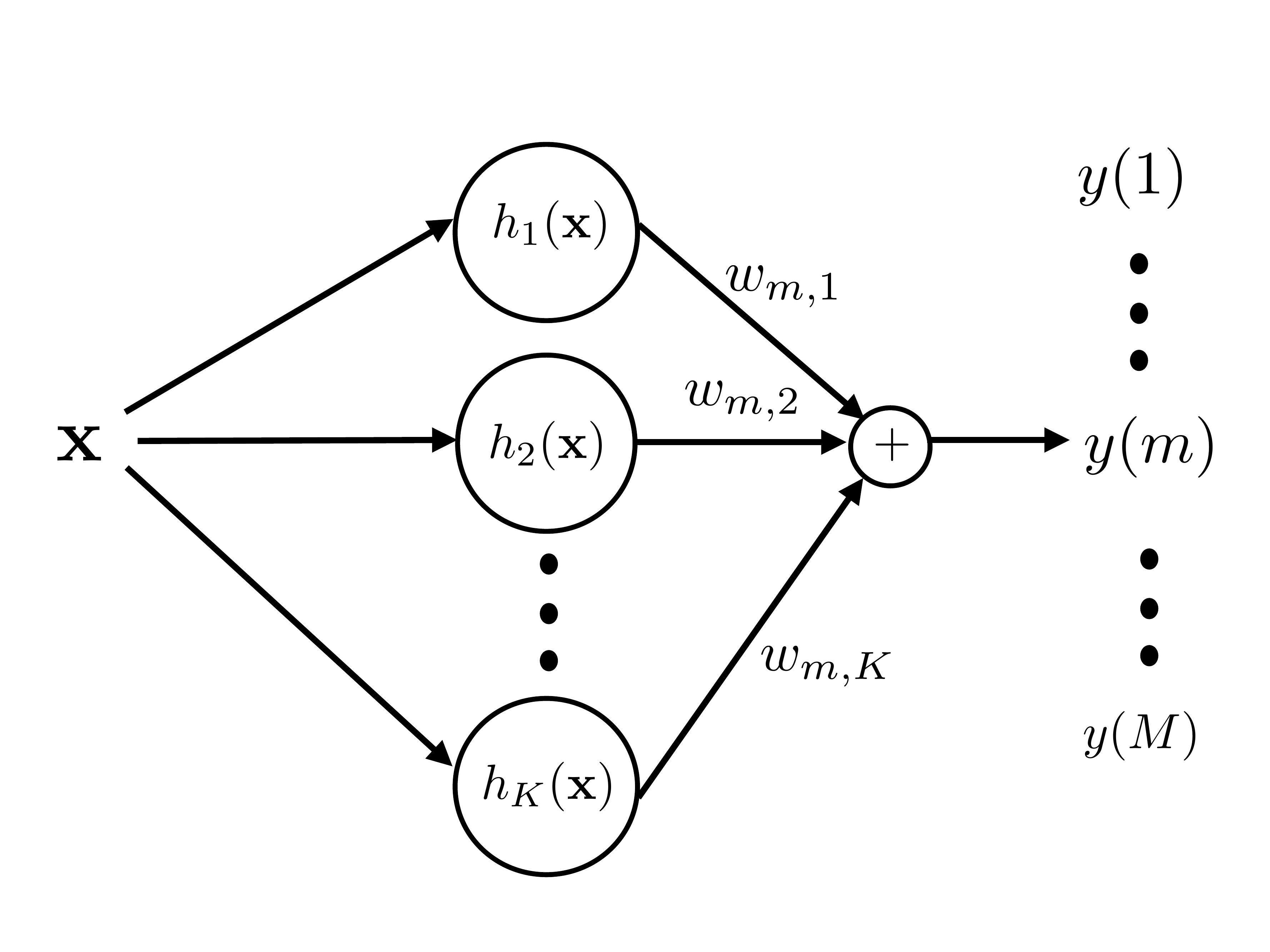}
	\caption{Schematic of an extreme learning machine for vector target.}
	\label{fig:elm_schematic}
	\vspace{-.1in}
\end{figure}
Consider a set of $N$ observations of input and target pairs $\{ (\mathbf{x}_n,\mathbf{y}_n) \}_{n=1}^N$. An extreme learning machine consists of the input layer, an output layer and a hidden layer of $K$ neurons as shown in Figure \ref{fig:elm_schematic}. The $k$th neuron implements a nonlinear operation $h_k(\cdot)$ on the input and is parameterized by set of variables $(\mathbf{a}_k,\mathbf{b}_k)$, that is,
\begin{equation*}
h_k(\mathbf{x})=G(\mathbf{x},\mathbf{a}_k,\mathbf{b}_k)
\end{equation*}
where $G(\cdot,\mathbf{a},\mathbf{b})$ is a parametric scalar function, for example, the sigmoid function $G(\mathbf{x},\mathbf{a},{b})=\displaystyle\frac{1}{1+\exp\left(-(\mathbf{a}^\top\mathbf{x}+b)\right)}$. The functions $h_k(\cdot)$ are referred to as the activation functions. The parameters $(\mathbf{a}_k,\mathbf{b}_k)$ are drawn randomly from a known probability distribution. The weights $w_{m,k}$ corresponds to the regression coefficient relating the output of the $k$th neuron to the $m$th component of the vector output $\mathbf{y}\in\mathbb{R}^M$. ELM models the output or target vector $\mathbf{y}$ as the output of linear regression such that
\begin{equation}
\mathbf{y}=\mathbf{W}^\top\mathbf{h}(\mathbf{x})
\end{equation}
where $\mathbf{h}(\mathbf{x})=[h_1(\mathbf{x}), h_2(\mathbf{x}), \cdots, h_K(\mathbf{x})]^\top$. 
In the ELM, we learn the regression matrix $\mathbf{W}$ by solving the following regularized least-squares problem:
\begin{equation}
\label{eq:elm_cost}
\arg\min_{\mathbf{W}}\sum_{n=1}^N \|\mathbf{t}_n-\mathbf{W}^\top\mathbf{h}(\mathbf{x}_n)\|_2^2 + \alpha \mbox{tr}(\mathbf{W}^\top\mathbf{W}),\,\,\alpha\geq0
\end{equation}
where $\mbox{tr}(\cdot)$ denotes the matrix trace operation.
Let us define the matrices $\T$ and $\H$ such that $
\T=[\t_1 \t_2 \cdots \t_N]^\top$ and $\H=[\h(\x_1) \h(\x_2) \cdots \h(\x_N)]^\top$.
Then, we have that
\begin{equation}
\label{eq:elm_cost2}
\mathbf{W}=\arg\min_{\mathbf{W}}\left( \|\T-\H\mathbf{W}\|_F^2 + \alpha \mbox{tr}(\mathbf{W}^\top\mathbf{W})\right),\,\,\alpha\geq0\nonumber
\end{equation}
where $\|\cdot\|_F$ denotes the Frobenius norm. The optimal value of $\mathbf{W}$ is obtained as \cite{elm_Huang2006}:
\begin{equation}
\label{eq:elm_w}
\mathbf{W}=(\mathbf{H}^\top\H+\alpha\mathbf{I}_N)^{-1}\mathbf{T}.
\end{equation}
Thus, ELM consists of input layer acted upon by activation functions with random parameters, followed by learning of the regression coefficients at the output layer.

\section{ELM over graph}
We next propose ELM for graph signal processing. We assume that the target signal $\mathbf{t}$ is a graph signal and may be corrupted by noise. Our goal is to learn an ELM such that the predicted output $\mathbf{y}$ is smooth over graph $\mathcal{G}$. We achieve this goal by enforcing a penalty of the form $\sum_nl(\mathbf{y}_n)$ while training the ELM. In other words, we learn the optimal regression coefficients as the minimizer of the following optimization problem:
\begin{align}
\label{eq:elm_cost_2}
C(\mathbf{W})=\sum_{n=1}^N \|\mathbf{t}_n-\mathbf{y}_n\|_2^2 + \alpha \mbox{tr}(\mathbf{W}^\top\mathbf{W})+\beta \sum_{n=1}^Nl( \mathbf{y}_n),\nonumber\\
\mbox{subject to  }\mathbf{y}_n=\mathbf{W}^\top\h(\x_n)\,\,\forall\,n,\quad\,\,\alpha,\beta\geq0.
\end{align}
We observe that \eqref{eq:elm_cost_2} is an optimization similar to that in \eqref{eq:elm_cost} and therefore represents an ELM with further regularization: the ELM output is enforced to be smooth over the graph $\mathcal{G}$.
Then, the predicted output for a new input $\mathbf{x}$ is given by 
\begin{equation*}
\label{eq: elm_g_output}
\mathbf{y}(\mathbf{x})=\mathbf{W}^\top_g\h(\x),
\end{equation*}
which we refer to as the output of the ELMG.
We next proceed to evaluate the optimal $\mathbf{W}_g$ matrix. Using properties of the matrix trace operation $\mbox{trace}(\cdot)$, we have that
\begin{align}
\label{eq:elm_g_cost_function_w}
C(\mathbf{W}) & =  \|\T-\H\mathbf{W}\|_F^2 + \alpha \mbox{tr}(\mathbf{W}^\top\mathbf{W})\nonumber\\
&\quad+\beta \,\sum_n \pmb\h(\mathbf{x}_n)^\top \mathbf{W L}\mathbf{W}^\top \pmb\h(\mathbf{x}_n) \nonumber\\
&= \mbox{tr}((\T-\H\mathbf{W})^\top(\T-\H\mathbf{W})) + \alpha \mbox{tr}(\mathbf{W}^\top\mathbf{W})\nonumber\\ 
&\quad+\beta \,\sum_n \pmb\h(\mathbf{x}_n)^\top \mathbf{W L}\mathbf{W}^\top \pmb\h(\mathbf{x}_n) \nonumber\\
& =  \mbox{tr}(\T^\top\T)-2\,\mbox{tr}\left(\mathbf{T}^\top\mathbf{\H}\mathbf{W} \right)  \nonumber \\
&\quad+\mbox{tr}\large(  \mathbf{W}^\top\mathbf{\H}^\top\mathbf{\H}\mathbf{W}\large) +\alpha\, \mbox{tr}(\mathbf{W}^\top\mathbf{W}) \nonumber \\
&\quad+  \beta \,\mbox{tr}\left( \displaystyle\sum_n \pmb\h(\mathbf{x}_n)^\top \mathbf{W L}\mathbf{W}^\top \pmb\h(\mathbf{x}_n)\right) \nonumber\\
&= 
\mbox{tr}(\T^\top\T)-2\,\mbox{tr}\left(\mathbf{T}^\top\mathbf{\H}\mathbf{W} \right) +\mbox{tr}\large(  \mathbf{W}^\top(\mathbf{\H}^\top\mathbf{\H}+\alpha\mathbf{I})\mathbf{W}\large) \nonumber\\&+  \beta\, \mbox{tr}\left(  \mathbf{W L}\mathbf{W}^\top \displaystyle\sum_n  \pmb\h(\mathbf{x}_n)\pmb\h(\mathbf{x}_n)^\top\right) \nonumber\\
& = \mbox{tr}(\T^\top\T)-2\,\mbox{tr}\left(\mathbf{T}^\top\mathbf{\H}\mathbf{W} \right) +\mbox{tr}\large(  \mathbf{W}^\top\mathbf{\H}^\top\mathbf{\H}\mathbf{W}\large)\nonumber \\
& \quad+ \alpha\, \mbox{tr}(\mathbf{W}^\top\mathbf{W}) +  \beta\, \mbox{tr}\left(  \mathbf{W}^\top\mathbf{\H}^\top\mathbf{\H}\mathbf{W}\mathbf{ L}\right).
\end{align}
$C(\mathbf{W})$ is quadratic in $\mathbf{W}$. Hence, we obtain the optimal and unique solution by setting the gradient of ${C}$ with respect to $\mathbf{W}$ equal to zero.
Using matrix derivative relations\cite{matrixbook}
\begin{eqnarray*}
	\begin{array}{rcl}
		\displaystyle\frac{\partial}{\partial \mathbf{W}} \mbox{tr}\left( \mathbf{M}_1 \mathbf{W} \right) & = & \mathbf{M}_1^{\top}, \\
		\displaystyle\frac{\partial}{\partial \mathbf{W}} \mbox{tr}\left( \mathbf{W}^{\top} \mathbf{M}_1 \mathbf{W} \mathbf{M}_2 \right) & = &  \mathbf{M}_1^{\top} \mathbf{W} \mathbf{M}_2^{\top} + \mathbf{M}_1 \mathbf{W} \mathbf{M}_2,
	\end{array}
\end{eqnarray*}
where $\mathbf{M}_1$ and $\mathbf{M}_2$ are matrices, 
and setting $\frac{\partial C}{\partial \mathbf{W}} =0$ we get that
\begin{eqnarray}
\begin{array}{l}
-\mathbf{\H}^\top\mathbf{T}+\mathbf{\H}^\top\mathbf{\H W} +\alpha \mathbf{W} +\beta \mathbf{\H}^\top\mathbf{\H W L}= \mathbf{0}, \\
\mathrm{or,} \,\,\,\, \left(\mathbf{\H}^\top\mathbf{\H}+\alpha \mathbf{I}_K\right)\mathbf{W}+\beta \mathbf{\H}^\top\mathbf{\H W L}=\mathbf{\H}^\top \mathbf{T}. \\    	 \end{array}
\label{eq:elm_g_J_ww}
\end{eqnarray}
On rearranging and vectorizing  terms in \eqref{eq:elm_g_J_ww}, we have that
\begin{align}
\mbox{vec}(\mathbf{\H}^\top\mathbf{T}) \! = \! \left[\!(\mathbf{I}_M \! \otimes \! (\!\mathbf{\H}^\top\mathbf{\H} \!+ \! \alpha\mathbf{I}_K) \!) \! + \! (\beta \mathbf{L} \! \otimes \! \mathbf{\H}^\top\mathbf{\H})\!\right] \! \mbox{vec}(\mathbf{W}), \nonumber 
\end{align}
where $\mbox{vec}(\cdot)$ denotes the standard vectorization operator and $\otimes$ denotes the Kronecker product operation \cite{Loan1}. Then, $\mathbf{W}_g$ follows the relation:
\begin{align}
\label{eq:elm_g_w}
\mbox{vec}({\mathbf{W}_g}) \! &= \!  \left[\!(\mathbf{I}_M \! \otimes \! (\!\mathbf{\H}^\top\mathbf{\H} \!+ \! \alpha\mathbf{I}_K) \!) \! + \! (\beta \mathbf{L} \! \otimes \! \mathbf{\H}^\top\mathbf{\H})\!\right]^{-1}   \mbox{vec}(\mathbf{\H}^\top\mathbf{T})\quad\nonumber\\
\! &= \!  \mathbf{D}^{-1}   (\mathbf{I}\otimes\mathbf{\H}^\top)\mbox{vec}(\mathbf{T})
\end{align}
where $\mathbf{D}=\mathbf{I}_M \! \otimes \! (\mathbf{\H}^\top\mathbf{\H}+\alpha\mathbf{I})+\beta \mathbf{L} \! \otimes \! \mathbf{\H}^\top\mathbf{\H}$.
We observe that on setting $\beta=0$ or $\mathbf{L}=\mathbf{0}$ which corresponds to having no graph, we get that
\begin{align*}
\mbox{vec}({\mathbf{W}_g}) \! &= \!  \left[\mathbf{I}_M \! \otimes \! (\!\mathbf{\H}^\top\mathbf{\H} \!+ \! \alpha\mathbf{I}_K)  \right]^{-1}   \mbox{vec}(\mathbf{\H}^\top\mathbf{T}),\,\,\mbox{or}\nonumber\\
\mathbf{W}_g&=(\mathbf{H}^\top\H+\alpha\mathbf{I}_N)^{-1}\mathbf{T}.
\end{align*}
In other words, ELMG reduces to the standard ELM when $\beta=0$.

\section{Smoothing effects for ELMG}
\label{smoothing}

We next discuss how the ELMG output for training data can be interpreted as a smoothing action across both observations and graph nodes. We show that the ELMG performs a shrinkage along the principal components of the graph Laplacian and kernel matrix. 
On vectorizing $\mathbf{Y}=\mathbf{H}\mathbf{W}_g$ and using \eqref{eq:elm_g_w}, we get that
\begin{align}
\mbox{vec}(\mathbf{Y}) &= (\mathbf{I}_M\otimes \mathbf{H})\mbox{vec}(\mathbf{W}_g) \nonumber\\
&=(\mathbf{I}_M\otimes \mathbf{H}) \mathbf{D}^{-1}(\mathbf{I}_M\otimes \mathbf{H}^\top)\mbox{vec}(\mathbf{T}).
\label{eq:vec_Y}	
\end{align}
Let $\mathbf{L}$ be diagonalizable with the eigendecomposition: 
\begin{eqnarray}
\mathbf{L}=\mathbf{V}\mathbf{J}_L\mathbf{V}^{\top},\,\,\,\,\nonumber
\end{eqnarray}
where 
$\mathbf{J}_L$ and $\mathbf{V}$ denote the eigenvalue and eigenvector matrices of $\mathbf{L}$, respectively, such that
\begin{eqnarray}
\mathbf{V}&=&[\mathbf{v}_1\,\mathbf{v}_2\cdots \mathbf{v}_N]\in\mathbb{R}^{M\times M}\nonumber\\
\mathbf{J}_L&=&\mbox{diag}(\lambda_1,\lambda_2,\cdots,\lambda_N)\in\mathbb{R}^{M\times M}.\nonumber
\end{eqnarray}
Further, let $\mathbf{H}=\mathbf{U}\mathbf{\Sigma}\mathbf{V}_H^\top$ denote the reduced singular-value decomposition of $\mathbf{H}$ such that 
\begin{eqnarray}
\mathbf{U}_H&=&[\mathbf{u}_{H,1}\,\mathbf{u}_{H,2}\cdots \mathbf{u}_{H,M}]\in\mathbb{R}^{M\times r}\,\,\nonumber\\ \mathbf{\Sigma}&=&\mbox{diag}(\sigma^2_1,\sigma^2_2,\cdots,\sigma^2_N)\in\mathbb{R}^{r\times r}\,\,  \nonumber\\
\mathbf{V}_H&=&[\mathbf{v}_{H,1}\,\mathbf{v}_{H,2}\cdots \mathbf{v}_{H,r}]\,\, \in\mathbb{R}^{r\times N}\mbox{and}\,\,\nonumber
\end{eqnarray}
where $r$ denotes the rank of $\mathbf{H}$ equal to the number of nonzero singular values, and $\mathbf{\Sigma}$ is the reduced singular value matrix which is the submatrix with only non-zero diagonal of the full singular value matrix.
Then, we have that
\begin{eqnarray}
\label{eq:krg_smooth}
\begin{array}{rl}
\mathbf{D}  & =  \mathbf{I}_M\otimes(\!\mathbf{\H}^\top\mathbf{\H}+\alpha\mathbf{I}) + \beta \mathbf{L} \! \otimes \! \mathbf{\H}^\top\mathbf{\H}\\
& =  [(\mathbf{V} \mathbf{I}_M \mathbf{V}^{\top}) \otimes (\mathbf{V}_H (\mathbf{\Sigma}^2 +\alpha\mathbf{I}_N) \mathbf{V}_H^{\top})]  \\
& \hspace{0.5cm} + [\beta(\mathbf{V}\mathbf{J}_L\mathbf{V}^{\top}) \otimes (\mathbf{V}_H\mathbf{\Sigma}^2\mathbf{V}_H^{\top})] \\
& \overset{(a)}{=}  [(\mathbf{V} \otimes \mathbf{V}_H) (\mathbf{I}_M \otimes (\mathbf{\Sigma}^2 +\alpha\mathbf{I}_N)) (\mathbf{V}^{\top} \otimes \mathbf{V}_H^{\top})] \\
& \hspace{0.5cm} + [\beta (\mathbf{V} \otimes \mathbf{V}_H) (\mathbf{J}_L \otimes \mathbf{\Sigma}^2) (\mathbf{V}^{\top} \otimes \mathbf{V}_H^{\top})] \\
& = (\mathbf{V} \otimes \mathbf{V}_H)  \mathbf{J} (\mathbf{V} \otimes \mathbf{V}_H) ^{\top},
\end{array}
\end{eqnarray}
where $\mathbf{J}=(\mathbf{I}_M \otimes (\mathbf{\Sigma}^2 +\alpha\mathbf{I}_N)) + \beta (\mathbf{J}_L \otimes \mathbf{\Sigma}^2)\in\mathbb{R}^{Mr\times Mr}$.
In (\ref{eq:krg_smooth})(a), we have used the distributivity of the Kronecker product: $(\mathbf{M}_1 \otimes \mathbf{M}_2) (\mathbf{M}_3 \otimes \mathbf{M}_4) = \mathbf{M}_1\mathbf{M}_3 \otimes \mathbf{M}_2 \mathbf{M}_4$ where $\{\mathbf{M}_i\}_{i=1}^{4}$ are four matrices.
Similarly,  we have that
\begin{eqnarray}
\label{eq:IkronH}
\begin{array}{rcl}
(\mathbf{I}_M \otimes \mathbf{H}) & = & (\mathbf{V} \mathbf{I}_M  \mathbf{V}^{\top}) \otimes (\mathbf{U}_H \mathbf{\Sigma} \mathbf{V}_H^{\top}) \\
& = &  (\mathbf{V} \otimes \mathbf{U}_H)   (\mathbf{I}_M \otimes \mathbf{\Sigma}) (\mathbf{V}^{\top} \otimes \mathbf{V}_H^{\top}) \\
\end{array}
\end{eqnarray}
Similarly,
\begin{eqnarray}
\label{eq:IkronH_2}
\begin{array}{rcl}
(\mathbf{I}_M \otimes \mathbf{H}^\top) & = & (\mathbf{V} \otimes \mathbf{V}_H)   (\mathbf{I}_M \otimes \mathbf{\Sigma}) (\mathbf{V}^{\top} \otimes \mathbf{U}_H^{\top}). 
\end{array}
\end{eqnarray}
Then, on substituting (\ref{eq:krg_smooth}), (\ref{eq:IkronH}), and (\ref{eq:IkronH_2}) in \eqref{eq:vec_Y}, we get that
\begin{align}
\mbox{vec}(\mathbf{Y}) 
& = (\mathbf{Z} (\mathbf{I}_M \otimes \mathbf{\Sigma})  \mathbf{J}^{-1} (\mathbf{I}_M \otimes \mathbf{\Sigma})  \mathbf{Z}^{\top})  \mbox{vec}(\mathbf{T})\nonumber\\
& = (\mathbf{Z} [(\mathbf{I}_M \otimes (\mathbf{I} +\alpha\mathbf{\Sigma}^{-2})) + \beta (\mathbf{J}_L \otimes \mathbf{I}))   ]^{-1}  \mathbf{Z}^{\top})  \mbox{vec}(\mathbf{T}),
\label{eq:vec_Y_simplified}	
\end{align}
where $\mathbf{Z} = \mathbf{V} \otimes \mathbf{U}_H$. Let $\mathbf{J}_F=[(\mathbf{I}_M \otimes (\mathbf{I} +\alpha\mathbf{\Sigma}^{-2})) + \beta (\mathbf{J}_L \otimes \mathbf{I}))   ]^{-1}$. Then, any diaognal element $\zeta_i$ of $\mathbf{J}_F$ is of the form
\begin{align*}
\zeta_i = [(1+\alpha\sigma_{i2}^{-2})+\beta\lambda_{i1}]^{-1}=\frac{1}{[(1+\beta\lambda_{i1})+\alpha/\sigma_{i2}^2]},\nonumber
\end{align*}
for some $i1\leq M$ and $i2\leq N$.
From \eqref{eq:vec_Y_simplified}, we have that 
\begin{eqnarray}
\label{eq:vec_y_pca}
\mbox{vec}\left(\mathbf{Y}\right)=\sum_{i=1}^{MN}\zeta_i\mathbf{z}_i\mathbf{z}_i^\top\mbox{vec}\left(\mathbf{T}\right),
\end{eqnarray}
where $\mathbf{z}_i$ are column vectors of $\mathbf{Z}$. \eqref{eq:vec_y_pca} expresses the prediction output $\mathbf{Y}$ as projections along the principal directions given by $\mathbf{z}_i=\mathbf{v}_{i1}\otimes\mathbf{u}_{H,i2}$ for some $i1\leq M$ and $i2\leq N$.
In the case when $\zeta_i\ll1$, the component in $\mbox{vec}(\mathbf{T})$ along $\mathbf{z}_i$ is effectively eliminated. The principal components corresponding to largest $\sigma^2$ correspond to the most informative directions. The components corresponding to smaller $\sigma^2$ are usually those of noise with high-frequency components. The eigenvectors of $\mathbf{L}$ corresponding to the smaller eigenvalues $\lambda$ are smooth over the graph \cite{Chung, Shuman}. We observe that the condition $\zeta_{i}\ll 1$ is achieved when corresponding $\sigma^2_{i1}$ is small and/ or $\lambda_{i2}$ is large. This corresponds to effectively retaining only components in $\mbox{vec}(\mathbf{T})$ which vary smoothly across the observation inputs $\{1,\cdots,N\}$ and and/ or smoothly varying across over the $M$ nodes of the graph. The extend of the smoothing achieved depends on the regularization parameters $\alpha$ and $\beta$. Thus, ELMG output corresponds to a smoothing or denoising operation over the training targets.

\section{Experiments}
We consider the application of the proposed ELM over graphs to two real-world graph signal datasets. In each of these datasets, the true targets $\mathbf{t}_{o,n}$'s are smooth graph signals which lie over a specified graph. We assume that the true target values are not observed and that we have access to only noisy target $\mathbf{t}_n$'s and the corresponding inputs $\mathbf{x}_n$'s. In order to simulate such a situation, we deliberately corrupt the true graph signal targets with additive white Gaussian noise $\mathbf{e}_n$:
\begin{equation*}
\mathbf{t}_n=\mathbf{t}_{o,n}+\mathbf{e}_n.
\end{equation*}
We then use the noisy targets for training  ELM and ELMG. The trained models are then used to predict targets for inputs from the test dataset. We consider the following three different activation functions popular in ELM literature:
\begin{enumerate}
	\item Sigmoid function: $G(\mathbf{x},\mathbf{a},{b})=\displaystyle\frac{1}{1+\exp(-(\mathbf{a}^\top\x+{b}))}$.
	\item Hardlimit function:
	$G(\mathbf{x},\mathbf{a},{b})=\begin{cases}
	1,\,\, \mbox{if}\,\, \mathbf{a}^\top\x+{b}\geq 0,\\
	0,\,\,\mbox{otherwise}.
	\end{cases}$
	\item Gaussian function: 
	$G(\mathbf{x},\mathbf{a},b)=\displaystyle\exp(-b\|\x-\mathbf{a}\|_2^2)$.
\end{enumerate}
The entries of parameters $\mathbf{a}$ and ${b}$ for each of the $K$ neurons are drawn independently from the standard normal distribution. The parameters $\alpha$ and $\beta$ are found by exhaustive grid search. 
We compare the prediction performance of both the strategies in terms of the normalized mean-square error (NMSE) for the test data, 
averaged over 100 different dataset partitions and noise realizations.
As discussed earlier, our hypothesis is that graph signal structure helps to improve prediction by the ELM. Our experiments below show that this is indeed the case\footnote{The codes used for experiments may be found at:\\
	https://www.kth.se/en/ees/omskolan/organisation/avdelningar/information-science-and-engineering/research/reproducibleresearch}.
\subsection{Temperature of cities in Sweden}

\begin{table}[t]
	\centering
	\begin{tabular}{ |c|c|c|c|c|c|c|c|c| } 
		\hline
		%
		$K$&ELM&ELMG&ELM&ELMG&ELM&ELMG\\
		\hline
		&\multicolumn{2}{c}{$N=4$}&\multicolumn{2}{|c}{$N=16$}&\multicolumn{2}{|c|}{$N=30$} \\
		\hline
		\multicolumn{7}{|c|}{Sigmoid function}\\
		\hline
		$10^2$&-3.74   &{\bf-6.58}	  	   &-7.72   &{\bf-9.36}	   &-7.87   &{\bf-8.99}\\
		$10^3$&-3.78   &{\bf-6.56}	  & -7.53   &{\bf-9.05}	   &-7.74   &{\bf-8.91}\\
		$10^4$&-3.94   &{\bf-6.71}	  	  & -7.63  & {\bf-9.24}   &-8.02  & {\bf-9.10}\\
		\hline
		\multicolumn{7}{|c|}{Hardlimit function}\\\hline
		$10^2$& -3.55   &{\bf-7.14}	   &-8.01   &{\bf-9.30}	  & -7.47   &{\bf-8.65}\\
		$10^3$&	-3.68   & {\bf-6.80}	  & -7.75   &{\bf-9.03}	   &-8.05   &{\bf-9.15}\\
		$10^4$&	-3.75   &{\bf-7.10}	   &-7.12  & {\bf-8.85}	   &-7.90  & {\bf-9.18}\\
		\hline
		\multicolumn{7}{|c|}{Gaussian function}\\
		\hline
		$10^2$&-3.73   &{\bf-6.34}	  	  & -7.61  & {\bf-9.11}	   &-7.77  & {\bf-8.92}\\
		$10^3$&-3.66   &{\bf-6.38}	   	   &-7.55  & {\bf-9.17}	  & -7.90   &{\bf-9.02}\\
		$10^4$&-3.70   &{\bf-6.69}	     &-7.58   &{\bf-9.09}	   &-7.80   &{\bf-9.00}\\
		\hline
	\end{tabular}
	\caption{NMSE for testing data as a function of $K$ and $N$ for temperature measurements.}
	\label{tab:elm_g_temp}
\end{table}
\begin{table}[t]
	\centering
	\begin{tabular}{ |c|c|c|c|c|c|c| } 
		\hline
		%
		$K$&ELM&ELMG&ELM&ELMG&ELM&ELMG\\
		\hline
		&\multicolumn{2}{c}{$N=10$}&\multicolumn{2}{|c}{$N=80$}&\multicolumn{2}{|c|}{$N=140$} \\
		\hline
		\multicolumn{7}{|c|}{Sigmoid function}\\
		\hline
		$10^2$&-14.99  &{\bf-19.24}	  	 & -23.31  &{\bf-24.26}	& -24.86&  {\bf-25.33}\\
		$10^3$&-14.95 & {\bf-19.13}&	  -23.38  &{\bf-24.26}&	-25.03  &{\bf-25.58}\\
		$10^4$&-14.84 & {\bf-19.26}	 &	-23.66 & {\bf-24.67}	&  -25.05 & {\bf-25.72}\\
		\hline
		\multicolumn{7}{|c|}{Hardlimit function}\\
		\hline
		$10^2$ &-14.99  &{\bf-19.24	} 	  &-23.31  &{\bf-24.26}	 &-24.87  &{\bf-25.33}\\
		$10^3$	&-14.95  &{\bf-19.13}		  &-23.38  &{\bf-24.26}	&-25.03  &{\bf-25.58}\\
		$10^4$&-14.84  &{\bf-19.26}	&-23.66  &{\bf-24.67}	  &-25.05 &{\bf-25.72}\\
		\hline
		\multicolumn{7}{|c|}{Gaussian function}\\
		\hline
		$10^2$ &-15.06 &{\bf-19.05}	  	&-23.55 & {\bf-24.62}&	  -25.68&  {\bf-26.14}\\
		$10^3$&-15.02  &{\bf-19.12}	   &-23.49 &{\bf-24.57}	  &-25.66  &{\bf-26.24}\\
		$10^4$&-15.09 &{\bf-19.13}	 	  &-23.56  &{\bf-24.63}	  &-25.75  &{\bf-26.30}\\
		\hline
	\end{tabular}
	\caption{NMSE for test set as a function of $K$ and $N$ for cerebellum fMRI data.}
	\label{tab:elm_g_cere}
\end{table}

\begin{figure}[t]
	\centering
	\includegraphics[width=2.7in]{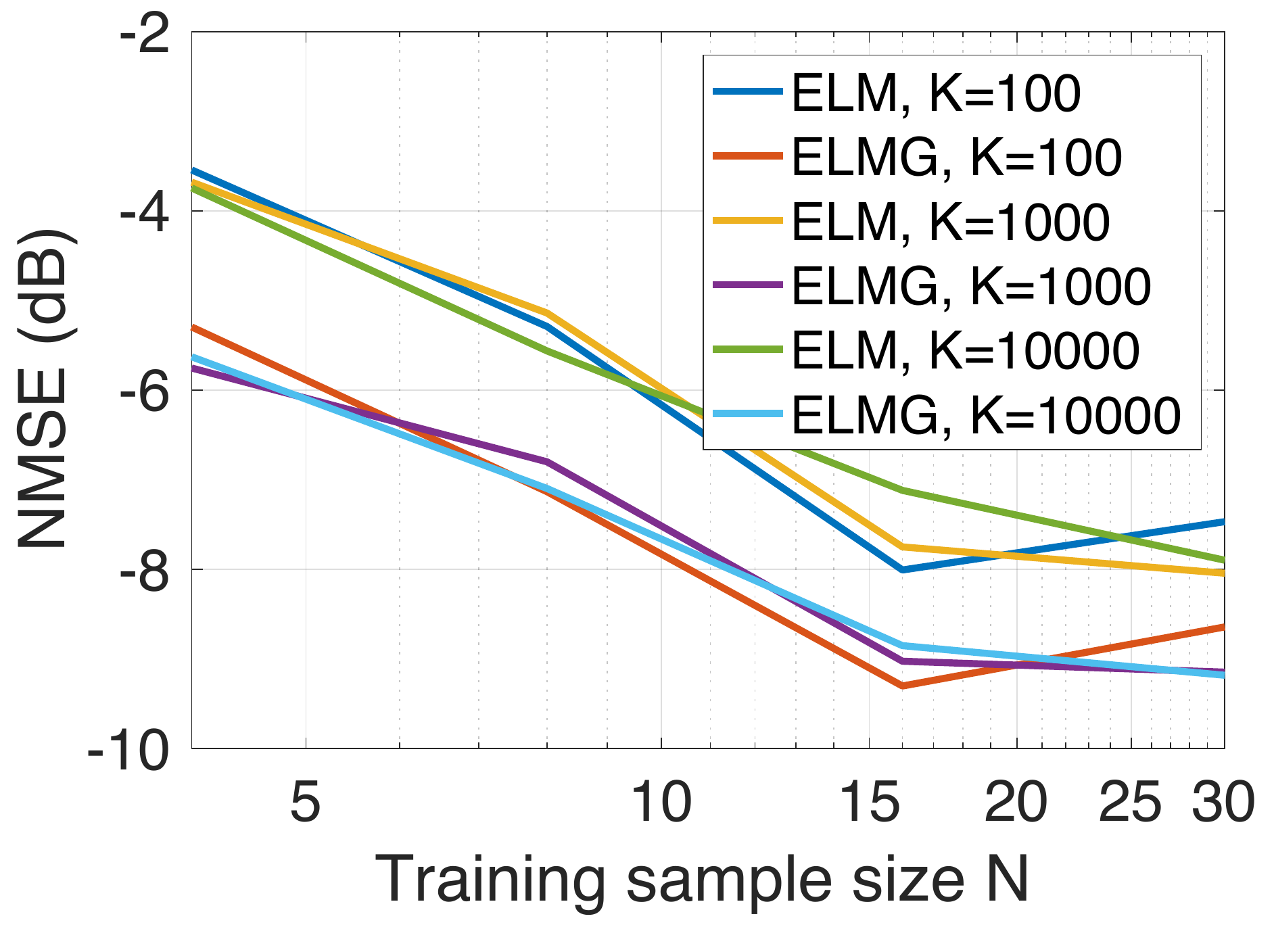}
	\caption{NMSE for temperature measurements with sigmoid activation function.}
	\label{fig:elm_g_temp}
\end{figure}

We consider the temperature measurements over the 45 largest cities in Sweden for the period of October to December 2015. We consider the geodesic graph whose adjacency matrix is given by
$a_{ij}=\exp{\left(-\frac{d_{ij}^2}{\sum_{i,j}d_{ij}^2}\right)}$, where $d_{ij}$ is the geodesic distance between the $i$th and $j$th cities. We consider the target to be the vector of  temperature measurements over all $45$ cities for a given day and the corresponding measurements from the previous day as the input $\mathbf{x}$. We construct noisy training targets by adding white Gaussian noise at a signal-to-noise (SNR) level of $5$dB. The NMSE as a function of $N$ for the sigmoid activation function is plotted in Figure \ref{fig:elm_g_temp}. We observe that the ELMG outperforms ELM by a signficant margin for all $K$, particularly at small sample sizes. As $N$ is increased, the performance of ELM and ELMG almost coincide. The NMSE obtained for different $N$ and $K$ values for all the three activation functions is listed in Table \ref{tab:elm_g_temp}.

\subsection{fMRI data for cerebellum}

\begin{figure}[t]
	\centering
	\includegraphics[width=2.7in]{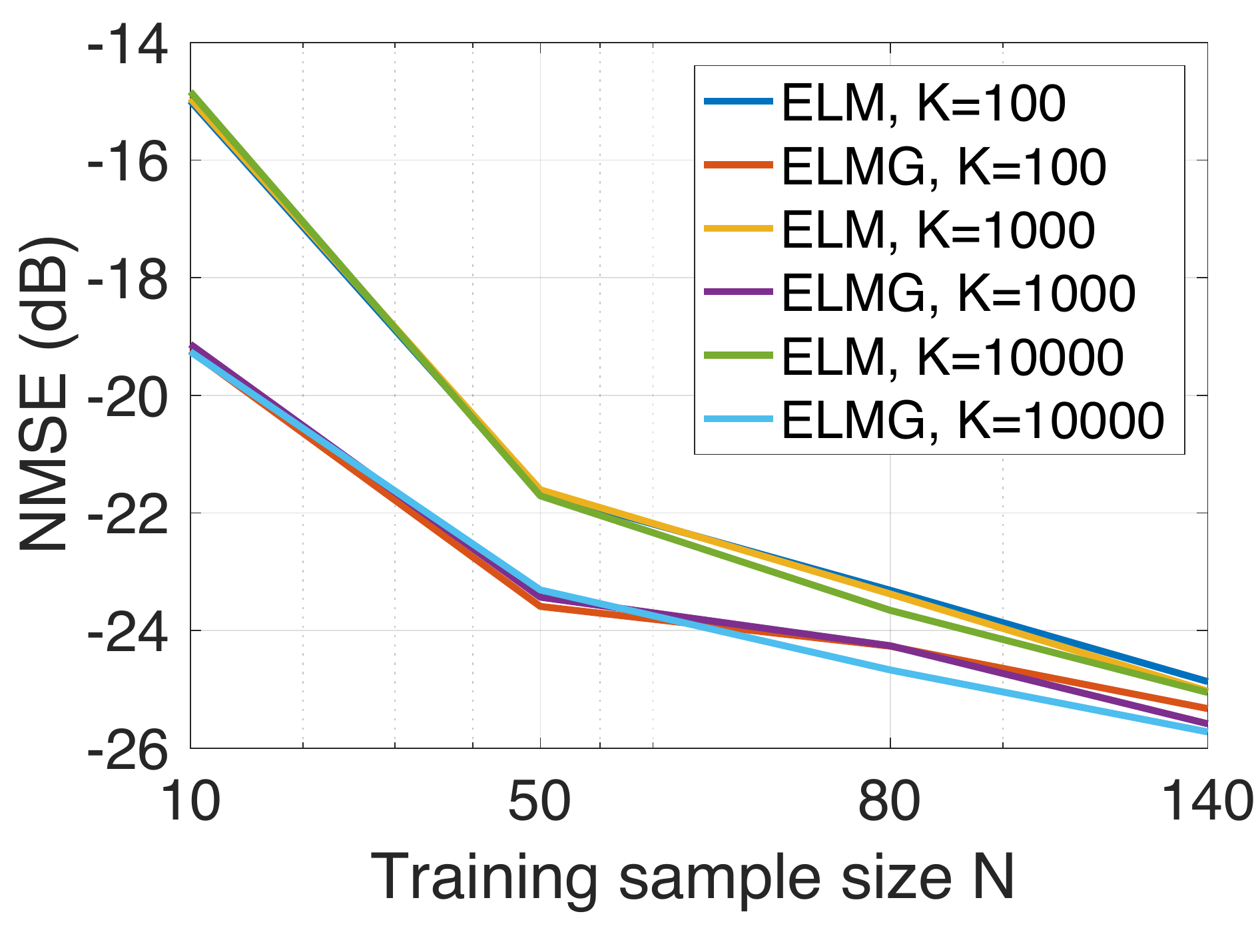}
	\caption{NMSE on for fMRI measurements with sigmoid activation function.}
	\label{fig:elm_g_cere}
\end{figure}

We next consider the functional magnetic resonance imaging (fMRI) data obtained for the cerebellum region of the brain used in \cite{Behjat_1}\footnote{The data is available publicly at https://openfmri.org/dataset/ds000102.}. The data consists of the intensity values at 4000 different voxels obtained from the fMRI of the cerebellum region. The graph is obtained by mapping the cerebellum voxels anatomically following the atlas template \cite{Behjat_cerebellum,cerebellum_atlas}. We refer to \cite{Behjat_1} for details of graph construction and associated signal extraction. We consider the first 1000 nodes in our analysis. Our goal is to use the intensity values at the first 100 vertices as input $\mathbf{x}\in\mathbb{R}^{100}$ and make predictions for the remaining 900 vertices, which forms the output $\mathbf{t}\in\mathbb{R}^{900}$. We have a total of 295 graph signals corresponding to different measurements from a single subject. We use a one half of the data for training and the other half for testing. As with the earlier experiment, we construct noisy training targets at a signal-to-noise (SNR) level of 5dB. The NMSE of the prediction mean for testing data, averaged over 100 different random partitions of the  dataset and different realizations of $\mathbf{a},\mathbf{b}$ and noise, for the three activation functions is listed in Table \ref{tab:elm_g_cere}. As also seen from Figure \ref{fig:elm_g_cere}, the trend remains the same as like the case of temperature data: ELMG outperforms ELM when the number of training samples is small, irrespective of the activation function and the number of neurons $K$. 

\section{Conclusions}
We bring together extreme learning machines and graph signal processing. Using the assumption that the signal to be predicted is smooth over a graph, the relevant regression problem uses graph-Laplacian matrix as well as a kernel between the observed inputs. The resulting solution has an interpretation of providing simultaneous smoothness across training samples and across graph nodes. Our hypothesis -- the graph knowledge will improve performance of extreme learning machine -- is verified by experiments on real data.

\bibliographystyle{IEEEbib}
\bibliography{refs,strings}

\end{document}